\documentclass{article}

\usepackage{arxiv}
\usepackage{amsmath} 
\usepackage{graphicx}
\usepackage[utf8]{inputenc} 
\usepackage[T1]{fontenc}    
\usepackage{hyperref}       
\usepackage{url}            
\usepackage{booktabs}       
\usepackage{amsfonts}       
\usepackage{nicefrac}       
\usepackage{microtype}      
\usepackage{lipsum}		
\usepackage{graphicx}
\usepackage{natbib}
\usepackage{doi}

\title{A Hybrid Random Forest and CNN Framework for Tile-Wise Oil-Water Classification in Hyperspectral Images}

\author{ \href{https://orcid.org/0000-0000-0000-0000}{\includegraphics[scale=0.06]{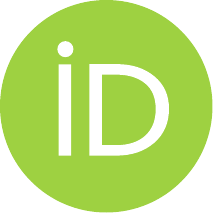}\hspace{1mm}Mehdi Nickzamir}\\
	Department of Computer Science\\
	Politecnico di Torino\\
	Turin, Italy \\
	\texttt{s323959@studenti.politoit} \\
	\And
	\href{https://orcid.org/0000-0000-0000-0000}{\includegraphics[scale=0.06]{orcid.pdf}\hspace{1mm}S. Mohammad Sheikh Ahmadi} \\
	Department of Computer Science\\
	Politecnico di Torino\\
	Turin, Italy \\
    \texttt{s327914@studenti.polito.it} \\
}

\renewcommand{\shorttitle}{\textit{arXiv} Template}

\hypersetup{
pdftitle={A template for the arxiv style},
pdfsubject={q-bio.NC, q-bio.QM},
pdfauthor={David S.~Hippocampus, Elias D.~Striatum},
pdfkeywords={First keyword, Second keyword, More},
}

\begin{document}
\maketitle

\begin{abstract}

This paper presents a novel hybrid Random Forest and Convolutional Neural Network (CNN) framework for oil-water classification in hyperspectral images (HSI). To address the challenge of preserving spatial context, we divided the images into smaller, non-overlapping tiles, which served as the basis for training, validation, and testing. Random Forest demonstrated strong performance in pixel-wise classification, outperforming models such as XGBoost, Attention-Based U-Net, and HybridSN. However, Random Forest loses spatial context, limiting its ability to fully exploit the spatial relationships in hyperspectral data. To improve performance, a CNN was trained on the probability maps generated by the Random Forest, leveraging the CNN’s capacity to incorporate spatial context. The hybrid approach achieved 7.6\% improvement in recall (to 0.85), 2.4\% improvement in F1 score (to 0.84), and 0.54\% improvement in AUC (to 0.99) compared to the baseline. These results highlight the effectiveness of combining probabilistic outputs with spatial feature learning for context-aware analysis of hyperspectral images.
\end{abstract}


\section{Introduction}
Oil spills have severe ecological, economic, and health consequences. In marine ecosystems, they threaten aquatic species, contaminate seafood, and disrupt the food chain, posing risks to human health. Additionally, oil spills impose financial and reputational damage on companies through fines and legal liabilities\cite{ElMoussaoui2023}. Timely detection is critical to mitigate these effects and prevent further damage.

Hyperspectral imaging (HSI) is a powerful tool for detecting oil spills, capturing detailed spectral information across hundreds of wavelengths. Unlike traditional RGB images, HSI identifies unique spectral “fingerprints” of oil spills, even when they are invisible to the naked eye. However, its effective use faces challenges, including high costs of airborne data collection, noise from environmental factors, and the need for robust models capable of handling high-dimensional data.

Data scarcity is another major hurdle, as obtaining labeled hyperspectral datasets is costly and time-intensive. While unsupervised methods avoid the need for labeled data, they often underperform compared to supervised techniques. Supervised methods like Random Forest excel at pixel-wise classification by leveraging spectral features but fail to incorporate spatial context, which is essential for producing coherent predictions across images.

To address these challenges, we propose a hybrid framework that combines Random Forest and Convolutional Neural Networks (CNNs). Random Forest handles pixel-wise classification, while CNNs capture spatial dependencies by refining the probabilistic outputs of the RF model. This combination balances spectral feature extraction and spatial learning, improving accuracy and consistency. Additionally, we adopt a tile-wise division of images to enhance model generalization. This approach enables faster and more reliable oil spill detection, contributing to environmentally conscious and efficient monitoring systems.

The implementation and dataset used in this study are publicly available on GitHub: [\href{https://github.com/adsp-polito/2024-P7-HSI/}{GitHub Repository}].

\section{Related Works}
Oil-water classification using hyperspectral images (HSI) has gained increasing attention due to its environmental significance, particularly in detecting and monitoring oil spills. This section focuses on recent advancements in HSI classification, with an emphasis on works closely aligned with the Hyperspectral Oil Spill Detection (HOSD) dataset, introduced by Puhong Duan.\cite{duan2023hosd}
\newline
The HOSD dataset, contains 18 hyperspectral images captured by the ARVIS sensor during the Deepwater Horizon oil spill. These images cover a spectral range of 365 nm to 2500 nm and are accompanied by reference maps that classify each pixel as either oil or non-oil. It is a valuable resource for advancing oil-water classification research.

Early work on the HOSD dataset explored unsupervised detection techniques for pixel level classification, such as the isolation forest (iForest)-based framework proposed by \cite{duan2023hosd}. This method employed a Gaussian statistical model to preprocess noisy spectral bands, followed by dimensionality reduction using kernel principal component analysis (KPCA). Probabilistic outputs from the iForest were refined using clustering algorithms and a support vector machine (SVM), achieving competitive image-wise classification accuracy. While effective, the approach did not integrate spatial relationships, which are critical for context-aware analysis in oil-water classification \cite{duan2023hosd}.

To address the limitations of pixel-wise methods, multiscale spectral-spatial learning frameworks have emerged as a promising direction for HSI classification. A notable contribution in this area is the multiscale spectral-spatial CNN (HyMSCN), which introduced an image-based classification framework designed to improve processing efficiency by integrating features from multiple receptive fields \cite{xu2021hymscn}. Unlike patch-based approaches, this method minimized redundancy in testing and effectively fused multiscale features to enhance classification accuracy. While the approach achieved strong performance on general-purpose hyperspectral datasets, it was not tailored for domain-specific datasets like HOSD, highlighting the need for specialized frameworks that address the unique spectral and spatial characteristics of oil spill imagery \cite{xu2021hymscn}.

Building upon the strengths of spectral-spatial integration, some efforts have emphasized the importance of contextual learning in HSI classification. The contextual CNN(HybridSN) proposed by \cite{zhang2020contextualcnn} demonstrated how a multi-scale convolutional filter bank could effectively exploit spatio-spectral relationships, producing a unified feature map for accurate pixel-wise classification. By leveraging deeper and wider architectures, this approach achieved high-ranking performance on standard datasets, such as Indian Pines and Salinas, underscoring the potential of contextual learning in enhancing classification accuracy \cite{zhang2020contextualcnn}.

While these approaches provide valuable insights, our method simplifies the training process by adopting a CNN on 2D probabilistic images generated by Random Forest rather than directly processing 3D hyperspectral data. This hybrid framework bridges the gap between probabilistic modeling and spatial feature learning, achieving superior oil-water classification performance on the HOSD dataset. By combining the strengths of Random Forest in both pixel-level and tile-wise classification with CNNs’ capacity to incorporate spatial context, the proposed approach offers a practical and efficient solution for context-aware analysis of hyperspectral images.

\section{Data Preprocessing}
\subsection{Dataset}
\begin{figure}[htbp]
    \centering
   \includegraphics[width=0.3\columnwidth]{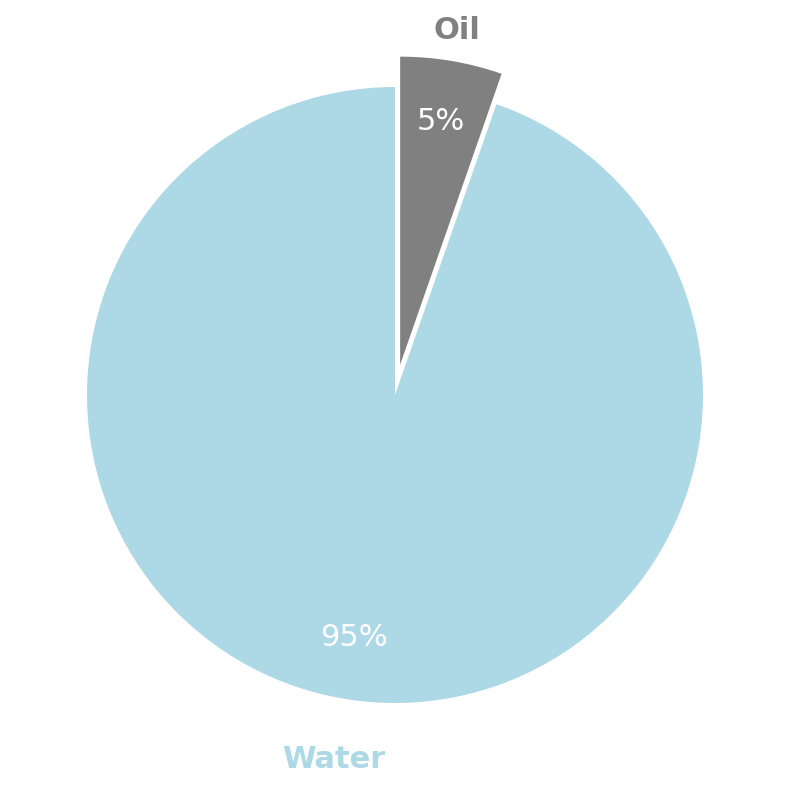}
    \caption{Distribution of Water and Oil Labels through the dataset.}
    \label{fig:Labels_Dis}
\end{figure}

The dataset used in this study is the publicly available Hyperspectral Oil Spill Detection (HOSD) dataset \cite{duan2023hosd}, which contains hyperspectral images documenting the Deepwater Horizon oil spill. The dataset has 18 images showing spectra from 365 nm to 2500 nm, labeled as oil or non-oil. The spatial resolution of the images varies due to different flight altitudes during data collection, as shown in Table \ref{tab:hosd_features}. The dataset's class imbalance (95\% water, 5\% oil) and the complexity of hyperspectral data present unique challenges, as depicted in Figure \ref{fig:Labels_Dis}.

\begin{table}[h]
    \centering
    \caption{Some features of the HOSD dataset.}
    \label{tab:hosd_features}
    \begin{tabular}{|l|l|l|l|}
        \hline
        \textbf{Data} & \textbf{Spatial Size} & \textbf{Resolution} & \textbf{Flight Time} \\ \hline
        GM1  & 1200$\times$633 & 7.6m & 5/17/2010 \\ \hline
        GM2  & 1881$\times$693 & 7.6m & 5/17/2010 \\ \hline
        GM3  & 1430$\times$691 & 7.6m & 5/17/2010 \\ \hline
        GM4  & 1700$\times$691 & 7.6m & 5/17/2010 \\ \hline
        GM5  & 2042$\times$673 & 7.6m & 5/17/2010 \\ \hline
        GM6  & 2128$\times$689 & 8.1m & 5/18/2010 \\ \hline
        GM7  & 2302$\times$479 & 3.3m & 7/09/2010 \\ \hline
        GM8  & 1668$\times$550 & 3.3m & 7/09/2010 \\ \hline
        GM9  & 1643$\times$447 & 3.2m & 7/09/2010 \\ \hline
        GM10 & 1110$\times$675 & 7.6m & 5/17/2010 \\ \hline
        GM11 & 1206$\times$675 & 7.6m & 5/17/2010 \\ \hline
        GM12 & 869$\times$649  & 7.6m & 5/06/2010 \\ \hline
        GM13 & 1135$\times$527 & 3.2m & 7/09/2010 \\ \hline
        GM14 & 1790$\times$527 & 3.2m & 7/09/2010 \\ \hline
        GM15 & 1777$\times$510 & 3.3m & 7/09/2010 \\ \hline
        GM16 & 1159$\times$388 & 3.2m & 7/09/2010 \\ \hline
        GM17 & 1136$\times$660 & 7.6m & 5/17/2010 \\ \hline
        GM18 & 1047$\times$550 & 3.3m & 7/09/2010 \\ \hline
    \end{tabular}
\end{table}

\begin{figure}[htbp]
    \centering
    \includegraphics[width=0.8\columnwidth]{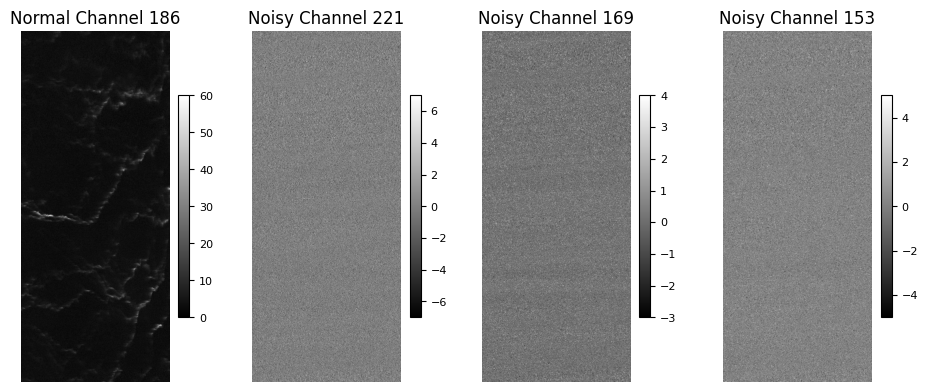}
    \caption{Visualization of three noisy channels and one normal channel in Image 14.}
    \label{fig:noise_normal_channels}
\end{figure}

\begin{figure*}[htbp]
    \centering
    \includegraphics[width=\textwidth]{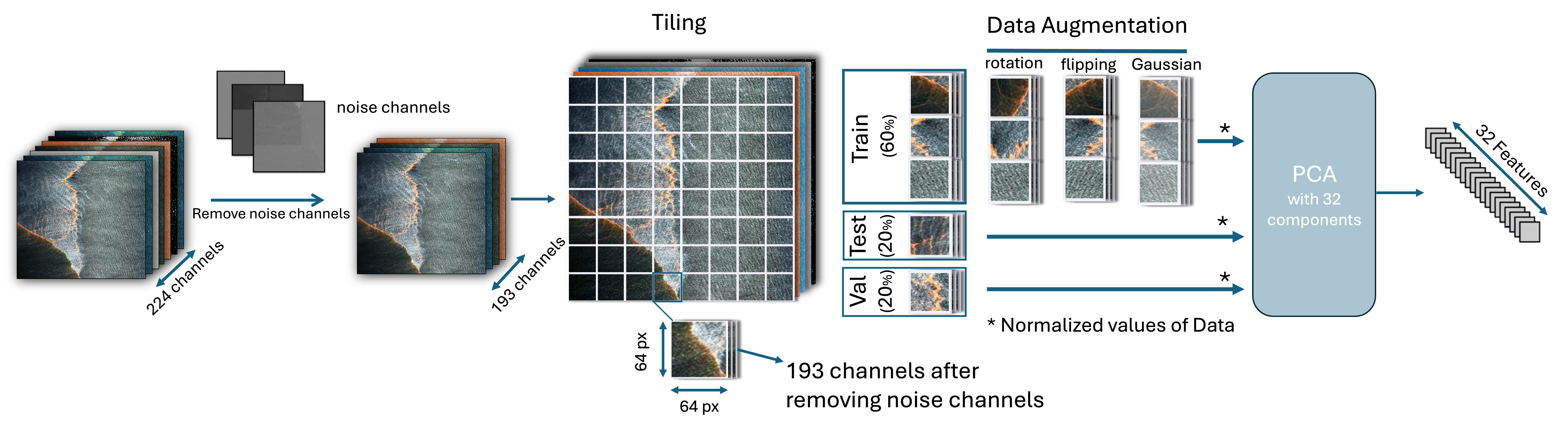}
    \caption{Pipeline of Data Preprocessing}
    \label{fig:PPP}
\end{figure*}

The preprocessing steps applied to address these challenges are summarized in the pipeline shown in Figure \ref{fig:PPP}. These steps include noisy channel removal, normalization, dimensionality reduction using PCA, tiling, and data augmentation. Each of these processes is detailed in the following subsections.

\subsection{Noisy Channel Removal and Normalization}

To improve dataset quality and facilitate dimensionality reduction, we carefully inspected each hyperspectral channel to detect and remove noisy ones. Noisy channels, characterized by low variance and lack of discernible patterns, fail to provide meaningful information for oil spill detection. Figure~\ref{fig:noise_normal_channels} illustrates examples of noisy and normal channels in the dataset.

To ensure consistency across the dataset, noisy channels were identified by intersecting channels marked as noisy across all images. This process resulted in the removal of 31 channels, as listed below: \newline
[103, 106, 107, 108, 109, 110, 111, 112, 113, 114, 152, 153, 154, 155, 156, 157, 158, 159, 160, 161, 162, 163, 164, 165, 166, 167, 168, 169, 221, 222, 223].

\begin{figure}[htbp]
    \centering
    \includegraphics[width=0.7\columnwidth]{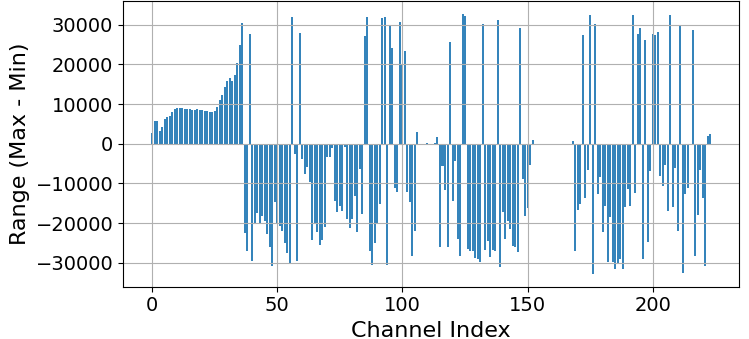}
    \caption{Range of Values in Data}
    \label{fig:Range}
\end{figure}

After removing noisy channels, normalization was applied to ensure that all channels contributed equally to variance. Figure~\ref{fig:Range} highlights the disparity in value ranges across channels, necessitating normalization. 

\vspace{\baselineskip}

\vspace{\baselineskip}
\vspace{\baselineskip}
\vspace{\baselineskip}
\vspace{\baselineskip}
\vspace{\baselineskip}

StandardScaler was used to transform each channel to have a mean of 0 and a standard deviation of 1:

\begin{equation}\label{eq:standard_normalization}
x_{\text{normalized}} = \frac{x - \mu}{\sigma}
\end{equation}

Where:
\begin{itemize}
\item $x$ is the original feature value,
\item $\mu$ is the mean of the feature values,
\item $\sigma$ is the standard deviation of the feature values.
\end{itemize}

Normalization ensured that Principal Component Analysis (PCA) focused on capturing meaningful patterns without being biased by differences in scale.

\begin{figure}[htbp]
    \centering
    \includegraphics[width=0.7\columnwidth]{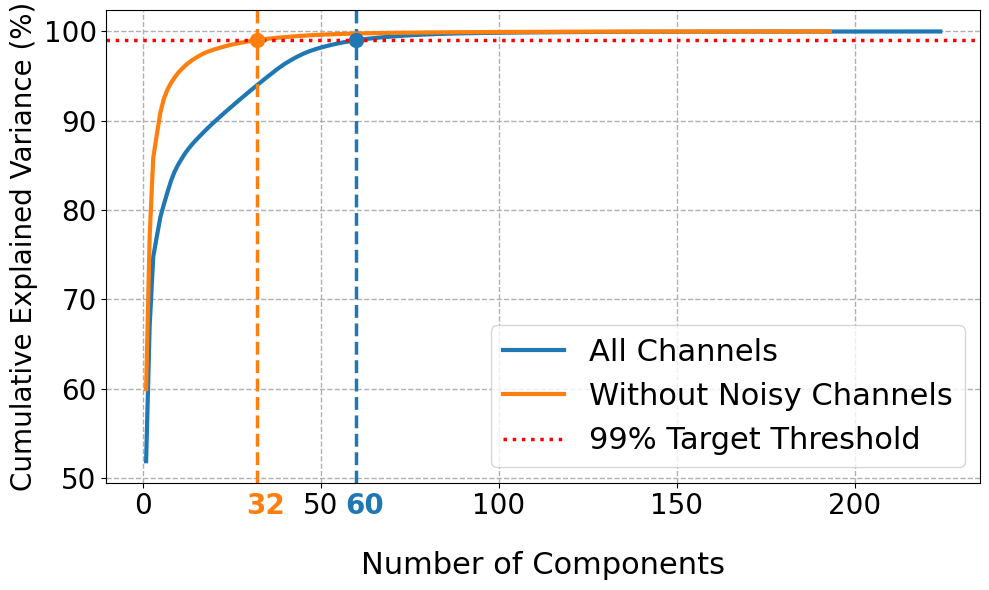}
    \caption{ Cumulative explained variance vs. PCA components }
    \label{fig:PCA32vsPCA60}
\end{figure}

\subsection{Dimensionality Reduction}
Although noisy channels were removed during preprocessing, the dataset remained high-dimensional due to the large number of spectral channels, posing challenges for computational efficiency and increasing the risk of overfitting.

To address this, Principal Component Analysis (PCA) was applied, reducing the dataset to 32 principal components while preserving 99\% of its variance, as shown in Figure \ref{fig:PCA32vsPCA60}. This reduction significantly simplified the dataset, retaining essential spectral information and improving its representation of meaningful features. Notably, the removal of noisy channels reduced the number of components required to explain 99\% of the variance from 60 to 32, highlighting the benefits of preprocessing.

\subsection{Tiling and Data Splitting into Train, Validation, and Test Sets}

To ensure robust training and evaluation, we divided each hyperspectral image into smaller, non-overlapping tiles of size 64*64, instead of assigning entire images to specific splits. This approach avoided overfitting, as the model was not confined to learning the context of some images while leaving others unrepresented.

After tiling, the tiles were shuffled, ensuring a fair distribution of data across the training, validation, and test sets. The shuffled tiles were then allocated as follows: 60\% for training, 20\% for validation, and 20\% for testing. This method distributed spatial contexts from different regions of the dataset into each subset. 

For edge tiles, which contained partially empty spaces due to the dimensions of the original images, we applied a padding strategy. These empty spaces were filled with a constant value calculated as the average reflectance of water for the corresponding channels in the affected tile. This ensured data consistency while preserving the integrity of the dataset.

\subsection{Data Augmentation}

To address the challenge of data scarcity, despite the dataset being relatively larger, in terms of number of images, compared to those used in similar studies, we applied effective data augmentation techniques: rotation, flipping, and Gaussian noise. The primary objective of these augmentations was to increase the diversity of the dataset while ensuring that no artificial patterns or anomalies were introduced, which could otherwise interfere with the model’s learning process.

An interesting and deliberate choice in our augmentation strategy was to exclude water-only tiles from augmentation. This decision was driven by the significant class imbalance in the dataset, where approximately 95\% of the labels represented water and only 5\% represented oil spills before augmentation. By focusing augmentation efforts solely on tiles containing oil spills, we aimed to improve the balance within the training set and enhance the model’s ability to recognize minority-class patterns. As a result, the proportion of oil tiles in the training set increased to approximately 10\%, compared to ~5\% in the validation and test sets, which remained representative of the real-world distribution.

The final dataset distribution of tiles is as follows:
\begin{itemize}
    \item \textbf{Training set:} 6,804 tiles
    \item \textbf{Validation set:} 759 tiles
    \item \textbf{Test set:} 780 tiles
\end{itemize}

\section{Methodology}
\subsection{Model Choice}
In designing our hybrid framework, we carefully selected two complementary models: Random Forest (RF) and Convolutional Neural Networks (CNNs), each addressing specific challenges of hyperspectral image (HSI) analysis.

Random Forest was chosen as the primary model due to its robustness in handling high-dimensional data and its ability to process noisy data effectively. Our dataset, despite being reduced is still high dimensional, and  RF’s ensemble-based architecture excels in isolating meaningful features from such complex data without requiring extensive parameter tuning. Additionally, RF performs well even with relatively small datasets, which is a critical factor given the scarcity of labeled HSI datasets.

Conversely, CNNs were incorporated to address a key limitation of Random Forest: the lack of spatial awareness. RF operates on individual pixels without considering relationships between neighboring pixels, which can lead to fragmented or inconsistent predictions. CNNs, with their convolutional layers, are adept at capturing spatial context, enabling the model to learn patterns and structures that span multiple pixels. This capability is particularly important for HSIs, where spatial features often complement spectral information in distinguishing oil spills from water.

\begin{figure*}[htbp]
    \centering
    \includegraphics[width=\textwidth]{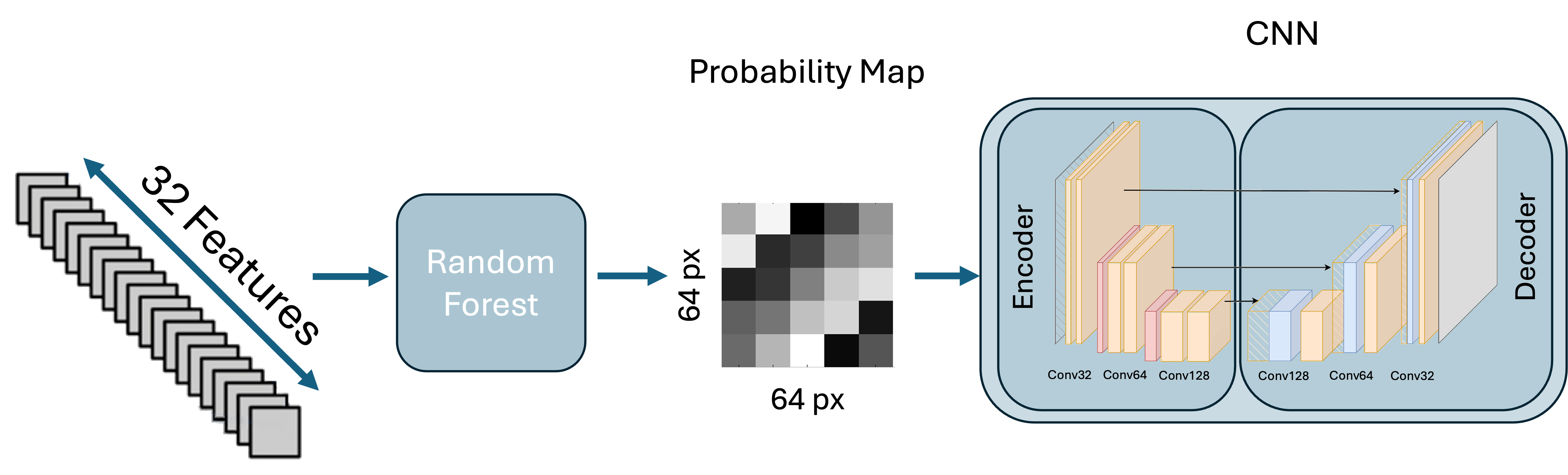}
    \caption{Overview of the Hybrid RF+CNN Framework for Oil Spill Detection}
    \label{fig:MP}
\end{figure*}

The hybrid framework leverages the strengths of both models, combining RF’s ability to handle high-dimensional noisy data with CNN’s capacity to refine predictions by incorporating spatial context. An overview of the hybrid framework is illustrated in Figure \ref{fig:MP}, which shows how the probabilistic outputs of RF are utilized as inputs for CNN to produce refined predictions.

\subsection{Training the Random Forest Model}

The  dataset was flattened to ensure compatibility with the Random Forest (RF) model. Each pixel, represented by 32 principal components was treated as an independent data point.The RF model was trained to perform pixel-wise classification, predicting the likelihood of each pixel belonging to the “oil” or “non-oil” class.  

Random Forest was configured with the following hyperparameters:
\begin{itemize}
\item \textbf{Number of Trees:} 100 — ensures a balance between model accuracy and computational efficiency.
\item \textbf{Random State:} 42 — ensures reproducibility of results.
\item \textbf{Max Features:} sqrt — uses the square root of the number of features for splitting, a commonly effective default.
\end{itemize}

the model achieved reliable results with the baseline hyperparameters, demonstrating its robustness.
However, due to resource constraints, we did not perform hyperparameter fine-tuning. The dataset’s size and memory requirements posed a challenge, as RF’s ensemble nature involves maintaining multiple decision trees in memory simultaneously.

\subsection{CNN Architecutre and Training}

\subsubsection{\textbf{Architecture}}
In the second phase of our hybrid framework, we employ a Convolutional Neural Network (CNN) to refine the probabilistic predictions generated by the Random Forest (RF) model. 

The CNN architecture consists of three main components:\newline
	\textbf{1.	Downsampling Path (Encoder)}: The encoder extracts hierarchical features from the input probability maps using convolutional layers with ReLU activations and same padding to retain spatial dimensions. Max-pooling is applied to progressively reduce spatial resolution, allowing the network to capture broader contextual information.\newline
	\textbf{2.	Upsampling Path (Decoder)}: The decoder restores the spatial resolution of feature maps using transposed convolutional layers. Skip connections directly link encoder and decoder features, enabling the integration of low-level spatial details with high-level abstractions. This enhances the refinement of predictions.\newline
	\textbf{3.	Output Layer}: A final convolutional layer with a sigmoid activation function produces a refined probability map, where each pixel represents the likelihood of belonging to the “oil” class.\newline

\subsubsection{\textbf{Training}}
The trained RF model generated probabilistic predictions for the validation and test sets, which were then reshaped into 64×64 tiles matching the CNN's input format. To ensure unbiased training, the validation set's prediction maps were split into a new training and validation set, with 80\% allocated for training and 20\% for validation.
This approach prevented the CNN from training on biased probability maps. Since the RF model had already seen the original training data, using its predictions from that set would have created overly optimistic maps too closely aligned with the ground truth. Instead, RF predictions on the validation set provided unbiased and realistic probability maps that were unseen during the RF model's training.
By training the CNN on these representative predictions, the model learned to refine its outputs effectively. The best-performing model was selected based on its validation AUC score. To mitigate overfitting, training was limited to 50 epochs due to the small number of tiles, which increased the risk of memorization.
\subsection{Metrics}

To assess the performance of our hybrid framework, we used five metrics: \textbf{Accuracy}, \textbf{Recall}, \textbf{Precision}, \textbf{F1-Score}, and \textbf{AUC-ROC}. However, due to the severe imbalance in the dataset (95\% water and 5\% oil), we prioritized \textbf{F1-Score} and \textbf{AUC} over accuracy, as the latter can be misleading in such scenarios. For example, a model predicting all labels as "water" would achieve a 95\% accuracy but fail entirely to detect oil.

\textbf{F1-Score}, the harmonic mean of precision and recall, was emphasized because it balances the trade-off between detecting oil pixels (recall) and minimizing false positives (precision). The \textbf{AUC-ROC}, which measures the model’s ability to distinguish between classes across various thresholds, complements F1-Score by providing an overall measure of classification performance.
\newline
The formulas for the key metrics are as follows:
\begin{equation}
    \text{Precision} = \frac{\text{True Positives (TP)}}{\text{True Positives (TP)} + \text{False Positives (FP)}}
\end{equation}
\begin{equation}
    \text{Recall} = \frac{\text{True Positives (TP)}}{\text{True Positives (TP)} + \text{False Negatives (FN)}}
\end{equation}
\begin{equation}
    \text{F1-Score} = 2 \cdot \frac{\text{Precision} \cdot \text{Recall}}{\text{Precision} + \text{Recall}}
\end{equation}
\begin{equation}
    \text{AUC} = \int_{0}^{1} TPR(FPR) \, d(FPR)
\end{equation}

Where:
\begin{itemize}
    \item TPR is the True Positive Rate, defined as $\frac{\text{TP}}{\text{TP} + \text{FN}}$.
    \item FPR is the False Positive Rate, defined as $\frac{\text{FP}}{\text{FP} + \text{TN}}$.
\end{itemize}

By focusing on F1-Score and AUC, we ensured that the evaluation reflected the model’s ability to balance precision and recall while effectively detecting the minority class (oil) in the highly imbalanced dataset.

\section{Experiments and Results}
In this section, we present the results of various models tested on the Hyperspectral Oil Spill Detection (HOSD) dataset. The primary goal of these experiments was to evaluate the performance of different methods, including Random Forest (RF), XGBoost (XGB), HybridSN, and Attention-based U-Net, and to compare them with our hybrid RF+CNN approach.

\subsection{Baseline and Comparison Models}
\textbf{Baseline Model (Random Forest):}  
Random Forest served as our baseline model due to its robustness in handling high-dimensional data and small datasets. It provided a strong starting point for pixel-wise classification, demonstrating high accuracy and acceptable F1 and AUC scores despite its lack of spatial awareness.
\newline
\textbf{XGBoost (XGB):}  
We tested XGBoost as an alternative ensemble learning method, but its performance was slightly lower than RF. XGB demonstrated effective feature handling but did not significantly outperform RF in terms of F1 and AUC.
\newline
\textbf{Attention-Based U-Net:}  
To evaluate whether a complex spatial-spectral model could perform better, we tested an attention-based U-Net. However, its performance was suboptimal due to the small dataset size, leading to overfitting and reduced generalization.
\newline
\textbf{HybridSN:}  
HybridSN, a spectral-spatial deep learning model proposed in prior researches, was also tested. While it outperformed the attention-based U-Net, it underperformed the Random Forest baseline model.
\newline
\textbf{Proposed Hybrid RF+CNN:}  
Our hybrid RF+CNN model was designed to combine the strengths of both RF and CNNs. This approach not only outperformed the baseline but also demonstrated improved consistency across tiles compared to the base model.

\subsection{Results}
The performance of all models is summarized in Table \ref{tab:model_performance}. While RF achieved competitive results, our hybrid RF+CNN approach outperformed it in key metrics such as F1 and AUC, demonstrating the effectiveness of incorporating spatial context.

\begin{table}[h]
\centering
\resizebox{0.7\columnwidth}{!}{ 
\begin{tabular}{|l|c|c|c|c|c|}
\hline
\textbf{Model Name} & \textbf{Accuracy} & \textbf{Precision} & \textbf{Recall} & \textbf{F1} & \textbf{AUC} \\ \hline
RF+CNN              & \textbf{0.9822}   & 0.8201             & \textbf{0.8519} & \textbf{0.8357} & \textbf{0.9940} \\ \hline
RF (Base Model)     & 0.9810            & \textbf{0.8409}    & 0.7921          & 0.8158      & 0.9886       \\ \hline
HybridSN            & 0.9792            & 0.7814             & 0.8443          & 0.8116      & 0.9919       \\ \hline
XGBoost (XGB)       & 0.9796            & 0.8264             & 0.7800          & 0.8025      & 0.9883       \\ \hline
Attention U-Net     & 0.9769            & 0.7912             & 0.7682          & 0.7795      & 0.9861       \\ \hline
\end{tabular}
}
\vspace{0.3cm}
\caption{Performance Comparison of Models (Sorted by F1-score)}
\label{tab:model_performance}
\end{table}

\subsection{Consistency Analysis}
\begin{figure}[h]
    \centering
    \includegraphics[width=0.7\columnwidth]{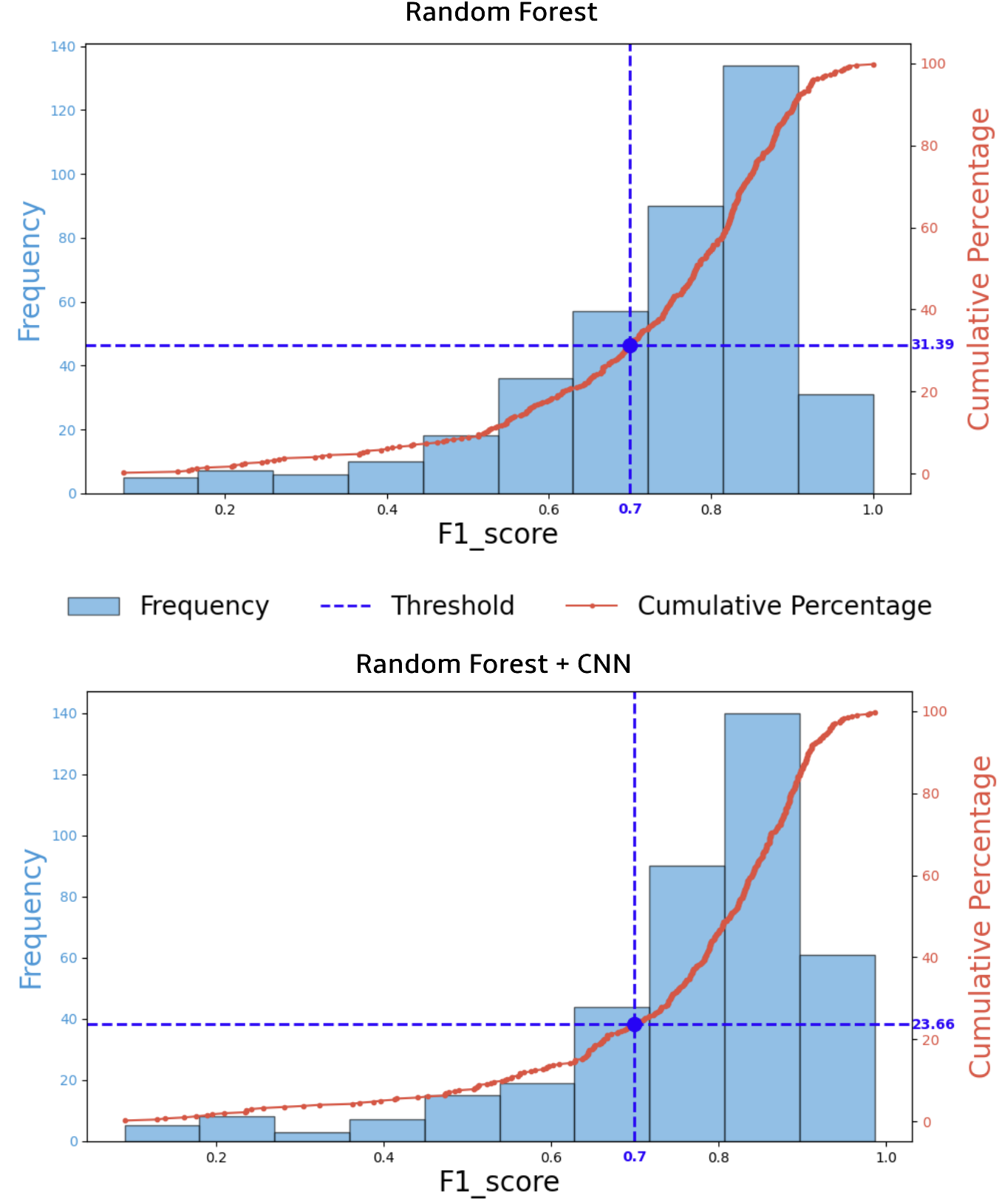}
    \caption{F1 score distributions across tiles for RF and RF+CNN.}
    \label{fig:f1_distribution}
\end{figure}

In addition to average performance metrics, we evaluated the consistency of predictions across tiles using F1 distributions. Analyzing tile-wise prediction histograms is important because average metrics, such as overall F1 score, can sometimes mask inconsistencies. A model may perform exceptionally well in certain tiles while failing in others, resulting in a high average performance but poor reliability in specific contexts. By examining the distribution of F1 scores, we gain deeper insights into how often the model struggles to make accurate predictions, enabling us to identify and address the contexts where the model is less effective.This analysis is particularly critical for hyperspectral oil spill detection, where environmental conditions and data quality vary widely across regions.
Figure \ref{fig:f1_distribution} illustrates the distribution of F1 scores for tiles predicted by the RF baseline and the RF+CNN hybrid approach.
\newline
\textbf{Baseline RF:}  
    Approximately 32\% of tiles achieved F1 scores below 0.7.
\newline
\textbf{Hybrid RF+CNN:}  
    With the hybrid approach, the proportion of tiles with F1 scores below 0.7 dropped to 24\%. 

The decrease in tiles with low F1 scores in the hybrid model shows how well RF’s ability to handle spectral features and CNN’s spatial context awareness work together. This improvement is vital for real-world applications, where consistent performance in different environmental conditions is key to reliable oil spill detection.

\section{Conclusion}
This study introduced a hybrid framework combining Random Forest (RF) and a Convolutional Neural Network (CNN) for hyperspectral oil spill detection. By leveraging RF’s robustness in handling high-dimensional, noisy data and CNN’s ability to capture spatial context, the framework effectively addressed the challenges of class imbalance and limited data.

Experimental results demonstrated that the hybrid RF+CNN model consistently outperformed other tested methods, including XGBoost, HybridSN, and attention-based U-Net, achieving higher F1-Score and AUC—key metrics for imbalanced datasets. Additionally, the hybrid model showed improved consistency across tiles, significantly reducing the proportion of low F1 scores compared to the RF baseline.

This framework highlights the potential of combining spectral and spatial features for effective hyperspectral image classification. Future work could explore the integration of advanced spatial-spectral techniques or domain-specific data augmentation strategies to further enhance the model’s robustness and generalization. Moreover, scaling this approach to larger datasets or real-time monitoring scenarios could unlock new possibilities for environmental applications and remote sensing.

\bibliographystyle{unsrtnat}
\bibliography{references}  






\end{document}